\documentclass[11pt,a4paper]{article}
\usepackage[hyperref]{emnlp2018}
\usepackage{times}
\usepackage{latexsym}
\usepackage[ruled,vlined,linesnumbered]{algorithm2e}
\usepackage{float}
\usepackage{graphicx}
\usepackage{lipsum}
\usepackage{url}
\usepackage{bm}
\usepackage{amssymb}
\usepackage{multicol}

\aclfinalcopy 

\title{Extending Neural Generative Conversational Model using External Knowledge Sources}

\author{Prasanna Parthasarathi \\
  McGill University, Canada\\
  {\tt pparth2@cs.mcgill.ca} \\\And
  Joelle Pineau \\
  McGill University, Canada\\
  Facebook AI Research, Canada \\
  {\tt jpineau@cs.mcgill.ca} \\}
\date{}

\begin{document}
\maketitle
\begin{abstract}
The use of connectionist approaches in conversational agents has been progressing rapidly due to the availability of large corpora. However current generative dialogue models often lack coherence and are content poor. This work proposes an architecture to incorporate unstructured knowledge sources to enhance the next utterance prediction in chit-chat type of generative dialogue models. We focus on Sequence-to-Sequence (Seq2Seq) conversational agents trained with the Reddit News dataset, and consider incorporating external knowledge from  Wikipedia summaries as well as from the NELL knowledge base. Our experiments show faster training time and improved perplexity when leveraging external knowledge. 
\end{abstract}

\section{Introduction}

Much of the research in dialogue systems from the last few years has focused on replacing all (or some) of its components with Deep Neural Network (DNN) architectures~\cite{pipelinedialogue,speechrec,li2017adversarial,serban2016building,serban2016hierarchical,neural_conv}. These DNN models are trained end-to-end with large corpora of human-to-human dialogues, and essentially learn to mimic human conversations.

Although these models can represent the input context, the need for a dedicated external memory to remember  information in context was pointed out and mechanisms were introduced in models like Memory Networks \cite{sukhbaatar2015end,bordes2016learning,gulcehre2018dynamic}, and the Neural Turing Machine \cite{graves2014neural}. Although these models, in theory, are better at maintaining the \emph{state} using their memory component, they require longer training time and excessive search for hyperparameters.

In this paper we explore the possibility of incorporating external information in dialogue systems as a mechanism to supplement the standard context encoding and facilitate the generation to be more specific with faster learning time. Furthermore, especially in the case of chit-chat systems, knowledge can be leveraged from different topics ( education, sports, news, travel, etc.).
Current memory-based architectures cannot efficiently handle access to large unstructured external knowledge sources.

In this work, we build on the popular Encoder-Decoder model, and incorporate external knowledge as an additional continuous vector. The proposed Extended Encoder-Decoder (Ext-ED) architecture learns to predict the embedding of the relevant external context during training and, during testing, uses an augmented \emph{state} of external context and encoder final state to generate the next utterance. We evaluate our model with experiments on Reddit News dataset, and consider using either the Wikipedia summary~\cite{scheepers2017compositionality} or the NELL KB~\cite{carlson2010toward} as a source of external context.

\section{Related Work}
\label{related}
Incorporating supplemental knowledge in neural conversational agents has been addressed in a few recent works on dialogue systems. Previous research however was mostly in the context of goal-driven dialogue systems, where the knowledge-base is highly structured, and queried to obtain very specific information (e.g. bus schedules, restaurant information, more broad tourist information).

Few goal-oriented dialogues research use external information directly from the web or relevant data sources. An exception is \cite{longenhanced}, which searches the web for relevant information that pertains to the input context and provides them as suggestions (like advising on places to visit while the user intends to visit a city). Similarly, instead of dynamically querying the web, \cite{ghazvininejad2017knowledge} pre-trains the model from facts in Foursquare \cite{foursquare} to select relevant facts as suggestions.

Moving closer to unstructured domains, but still within task-driven conversations, \cite{lowe2015incorporating} proposes a way of retrieving relevant responses based on external knowledge sources. The model selects relevant knowledge from Ubuntu man pages and uses it to retrieve a relevant context-response pair that is inline with the knowledge extracted. Similarly, \cite{young2017augmenting} extracts relations within the context and parses over it to score the message response pairs. The relational knowledge provides a way of incorporating useful knowledge or \emph{common sense} as termed by the authors.

Though \cite{guu2017generating} did not directly make use of an external knowledge source, they used an \emph{edit vector} that aids in editing a sampled prototype sentence. This is relevant to our proposed model as the generated response is conditioned on a supplementary vector similar to the external context vector discussed later in this paper.

\section{Technical Background}
\label{back}
\subsection{Recurrent Neural Networks}
The Recurrent Neural Network (RNN) is a variant of neural network used for learning representations of inputs, $\bm{x_{1:T}}$, that have an inherent sequential structure (speech, video, dialogue etc.). In natural language processing, RNNs are used to learn language models that generalize over n-gram models \cite{katzNgram}. The RNN maintains a hidden state, $h_t$, that is an abstraction of inputs observed until time-step $t$ of the input sequence, and uses $x_t$ to operate on them. RNN uses two projection functions, $U$ and $W$, for computing operations on input and hidden states respectively. A third function, $V$, to map $h_t$ to the output, $y_t$, the output of the RNN at every time step $t$. $y_t$, is a distribution over the next token given the previous tokens, and is computed as a function of $h_t$. The functions of the RNN can be explained as shown in Equations \ref{state} and \ref{output},

\begin{equation}
h_t = \emph{g}\left(U\cdot x_t + W\cdot h_{t-1} + b\right),
\label{state}
\end{equation}
\begin{equation}
y_t = V\cdot h_t + d,
\label{output}
\end{equation}
where $y_t$ is the output, $x_t$ is the vector representation of input token, $h_t$ is the internal state of the RNN at time $t$ and $g$ is a non-linear function (like \emph{tanh} or \emph{sigmoid}). RNNs are trained with Back Propagation Through Time (BPTT) \cite{rumelhart} to compute weight updates using the derivative of a loss function with respect to the parameters over all previous time-steps. 

\subsection{Seq2Seq Dialogue Architecture}

Generative dialogue models \cite{seq2seq,serban2015survey} extends the language model learned by RNNs to generate natural language that are conditioned not only on the previous words generated in the response but also on a representation of the input context. The ability of such a learning module to \emph{understand} an input sequence of words (that we call context ($\bm{c_{1:T}^i}$)) and \emph{generate} a response $\bm{r_{1:T}^i}$ tantamount to solving the \textit{dialogue task}. 

\cite{neural_conv} first proposed a vanilla LSTM \cite{lstm} dialogue model that encodes a given context with an LSTM (\emph{Encoder}) and feeds it to another LSTM (\emph{Decoder}) that generates a response token-by-token. Here the choice of encoder and decoder modules can be any recurrent architectures like GRU \cite{cho2014gru}, RNN, Bi-directional LSTM \cite{bidirectional}, etc. The model is then trained to learn a conditional distribution over the vocabulary for generating the next token in response to the $i^{th}$ context as shown in Equation \ref{generative}:
\begin{equation}
P\left(\bm{r_{1:T}^i}\mid \bm{c_{1:T}^i}\right) = \Pi_{k=1}^TP\left(r^i_k\mid \bm{r_{1:k-1}^i},\bm{c_{1:T}^i}\right).
\label{generative}
\end{equation}

With neural language models, this form can aid in maintaining long term dependencies and the next word distribution can be made to conditionally depend on an arbitrarily long context. Sophisticated models have made significant architectural improvements to aid better modelling of the contexts \cite{serban2016hierarchical}.

\section{Extended Encoder Decoder Architecture}
\label{model}
The primary objective of the proposed architecture is to supplement response generation with external knowledge relevant to the context. Most of the knowledge sources that are available are free-form and lack suitable structure for easy querying of relevant information. In this work, we attempt to incorporate such unstructured knowledge corpora for dialogue generation in Seq2Seq models.

\begin{figure}[H]
    \centering
    \includegraphics[width=0.55\textwidth]{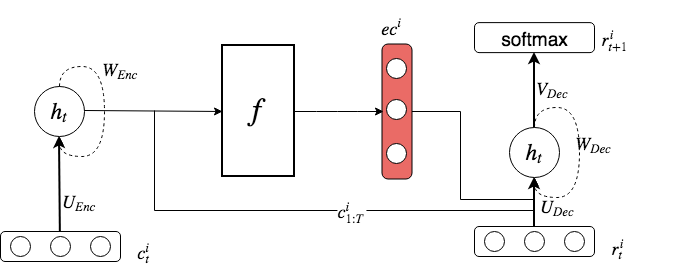}
    \caption{Architecture Diagram of Extended Encoder Decoder Model.}
    \label{eed}
\end{figure} 

\subsection{The Model}
\label{modeldefinition}

The Extended Encoder Decoder (Ext-ED) model, shown in Figure \ref{eed}, uses an encoder LSTM (parameterized as $\bm{\Theta}_{Enc}$) to encode the $i^{th}$ context, $\bm{c_{1:T}^i}$, and a fully connected layer ($f$) to predict an external context vector ($\bm{ec^i}$) conditioned on the encoded context. The predicted external context vector, $\bm{ec^i}$, is provided to the decoder LSTM ($\bm{\Theta}_{Dec}$) at every step, augmented with the encoder final state and previous predicted token, to generate the next token in the response:
\begin{equation}
    P\left(\bm{ec^i} \mid \bm{c_{1:T}^i}\right) = f\left(\bm{\Theta_{Enc}\left(c_{1:T}^i\right)}\right)
    \label{external}
\end{equation}
\begin{equation}
    P\left(\bm{r_{1:T}^i}\mid \bm{c_{1:T}^i}\right) = \Pi_{k=1}^TP\left(r_k^i\mid \bm{r_{1:k-1}^i},\bm{c_{1:T}^i}, \bm{ec^i}\right).
    \label{decodereqn}
\end{equation}

The decoder is provided with an encoding of the context along with the external knowledge encoding, as $\bm{ec^i}$ acts as information supplement to the knowledge available in the context as shown in Equations \ref{external} and \ref{decodereqn}. 

During training, the gradients for Ext-ED parameters ($f,\Theta_{Enc}, \Theta_{Dec}$) are computed by backpropagating the gradients for parameters with respect to losses in Equations \ref{loss1}, \ref{loss2} and \ref{loss3}:

\begin{equation}
    \mathcal{L}_1 = \sum_{k=1}^T Q\left(r^i \mid \hat{ec}^i, c^i\right) \log P\left(r^i \mid c^i\right),
    \label{loss1}
\end{equation}
\begin{equation}
    \mathcal{L}_2 = \Vert \hat{\bm{ec}}^i - \bm{ec^i} \Vert_2,
    \label{loss2}
\end{equation}
\begin{equation}
    \mathcal{L}_3 = - \sum_{k=1}^T Q\left(r^i \mid c^i, \bm{0}\right) \log P\left(r^i \mid c^i\right).
    \label{loss3}
\end{equation}

 Here $\mathcal{L}_1$ is the log-likelihood that is used to make the model distribution mimic the data distribution. $\mathcal{L}_2$ trains $f$ to correctly predict $ec^i$, and $\mathcal{L}_3$ trains $\Theta_{Dec}$ to make use of the external context by forcing it the model distribution to diverge when not provided with the external context ($ec^i$ is set to ${\bm 0}$ vector). In the loss equations, \emph{P} and \emph{Q} represent the data and the model learned distributions respectively, and $\bm{ec^i}$ and $\bm{\hat{ec}^i}$ represent true and model($f$) predicted external knowledge encoding.

\subsection{External Context Vector}
\label{modeldetails}
We use Wikipedia summary \cite{scheepers2017compositionality} and NELL knowledge base (KB)  \cite{carlson2010toward} to compute the external knowledge encoding for every context in the context-response pairs. Algorithm \ref{extcontalg} oultines the pseudocode for computing the external context vector ($\bm{ec^i}$). For $i^{th}$ input context, the methods \emph{Return\_All\_Values\_for\_Entity} or  \emph{Wiki\_Summary\_Query} is used to extract the external knowledge vector, $ec^i$, from NELL KB or Wikipedia summary sources.

\SetAlFnt{\small\sf}
   \begin{algorithm}[h]
    \label{extcontalg}
    \caption{Get\_External\_Context \_Vector ($\bm{c^i_{1:T}}$)}
    
    \SetKw{return}{return}
    \SetKw{from}{from}
    \SetKw{downto}{downto}
    \SetKw{step}{step}
    
    $\bm{ec^i}$ $\gets$ zero\_vector \\
    $\#\_Ext\_Tokens$ $\gets$ 0 \\
   \For{t in range (1,T)}{
    \If{$\bm{c_t^i}$ is not a \textbf{stop word}}{
    External\_Tokens$_{List}$ $\gets$ Wiki\_Summary\_Query($\bm{c_t^i}$) \\ \% Return\_All\_Values\_for\_Entity($\bm{c_t^i}$)\\
    \For{token in External\_Tokens$_{List}$}{
    $\bm{ec^i}$ $\gets$ $\bm{ec^i}$ + GloVe\_Embedding(token)\\
    $\#\_Ext\_Tokens$ $\gets$ $\#\_Ext\_Tokens$ + 1
    }
    }
}
        return $\frac{\bm{ec^i}}{\#\_Ext\_Tokens}$
        \end{algorithm}

The external context encoding, $\bm{ec^i}$, is a fixed length continuous embedding of the knowledge from external sources, as having all the words sampled (represented as a Bag of Words ) proved to be a severe computational overhead because of sparsity.

The continuous embedding of external context provides an additional conditioning with relevant external knowledge that is used to generate the next utterance. Although this is the intended hypothesis, there are certain expected characteristics that are desirable of external knowledge sources for them to be useful:
\begin{itemize}
    \item The knowledge vectors being away from the mean of their distribution.
    \item The knowledge vectors having high variance.
\end{itemize}

We analyzed the knowledge vectors constructed using the two knowledge sources and the distribution of distances of $\bm{ec^i}$ from their mean. The variance of this distribution in NELL KB was \emph{1.44} and from that of Wikipedia summary was \emph{0.73}. Mean distance was very low (\emph{0.77}) in the case of Wikipedia compared to that of NELL (\emph{2.16}). We observed that the vectors not being spread out made them less useful than the encoded context itself in the initial experiments. 

\section{Experiments and Discussion}
\label{exper}
We evaluated Ext-ED by training with Reddit News dataset, and incorporated Wikipedia Summary and NELL KB as sources of external knowledge. The objective of the experiments are three-fold: (1) to evaluate the ability of the model to make use of the external context to condition the response; (2) to analyze the training time with the additional knowledge provided; and (3) to observe any tangible differences in training with the two knowledge sources.

\begin{figure}
    \centering
    \includegraphics[width=0.5\textwidth]{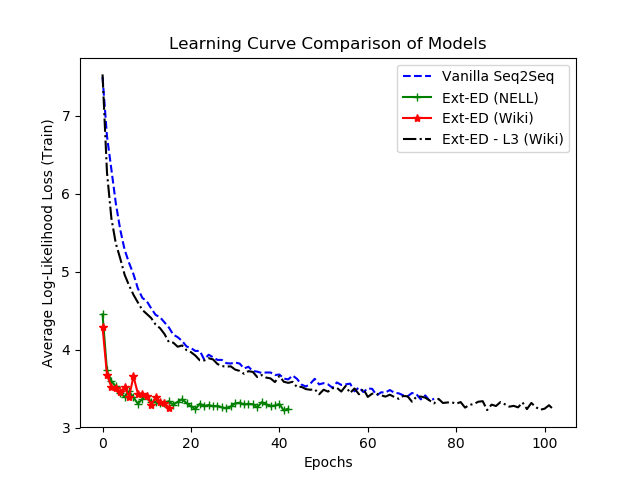}
    \caption{Convergence of Sequence loss (cross-entropy loss in sequential outputs) over different models during training.(-L3 in legend denote exclusion of $\mathcal{L}_3$ loss from gradient computation.)}
   \label{train}
\end{figure}

For the first analysis, we trained Ext-ED with external context ($\mathbb{R}^{100}$) and validated it without providing it (see Ext-ED - $\mathcal{L}_3$ Ablation in Table \ref{scores}). Without the inclusion of $\mathcal{L}_3$, Ext-ED did not find the external context useful and the performance was not very different from a Vanilla Seq2Seq dialogue model (Figure \ref{train}). The encoder context had enough variance to be a viable information source and hence the external context was ignored. This can be observed from similar learning curves of \emph{Vanilla Seq2Seq, Ext-ED - L3 (Wiki)} models in Figure \ref{train}.

With propagating back gradients with respect to $\mathcal{L}_3$, we observed that the model learns to use the external context, but, as discussed in Section \ref{modeldetails}, the variance in the external context vectors constructed using the two knowledge sources was too low. To fix this, we scaled the external context vectors with $\mathcal{N}(4,1)$. This improved the variance in the knowledge that subsequently improved the usefulness of these vectors which was also observed in Figure \ref{train}.

\begin{table}[h]
    \centering
    \begin{tabular}{|p{3.5cm}|p{1cm}|p{1.4cm}|}
        \hline
         \textbf{Model} & \textbf{PPL} & \textbf{BLEU-4}  \\
        \hline
         Vanilla Seq2Seq & 38.09 & 0.437 \\
         Ext-ED - $\mathcal{L}_3$ (Wiki) & 38.37 & 0.435 \\
         Ext-ED - $\mathcal{L}_3$ Ablation & 37.06 & 0.425 \\
         Ext-ED (Wiki) & {\bf 30.26} & {\bf 0.53} \\
         Ext-ED (NELL KB) & {\bf 29.07} & {\bf 0.525}\\
         Ext-ED Ablation & 601.8 & 0.274 \\
         \hline
    \end{tabular}
    \caption{Comparison of BLEU (sentence\_bleu) and Perplexity scores on validation set across different models.}
    \label{scores}
\end{table}

The provision of external knowledge improved the Perplexity and BLEU-4 scores as shown in Table \ref{scores}. Though the improvements are reasonable, the metrics used are not strong indicators for evaluating the influence of external contexts in dialogue. But, they do indicate that the prediction accuracy is improved with the inclusion of external knowledge sources. The poor perplexity for Ext-ED Ablation is because the model is conditioned to predict the next utterance  using $\bm{ec^i}$ and when not provided the context alone is not sufficiently informative. Another way to interpret this would be to see that the external context and and the dialogue context provide complementary information for better predicting the next utterance. Further, the experiments illustrated that the model, when provided with an \emph{informative} source of knowledge (the one that has higher variance), will let the model converge faster. One possible hypothesis is that $\bm{ec^i}$, which has high variance and is provided as input in every step of decoding, is learned before the RNN parameters converge. The information in $\bm{ec^i}$ is relevant to the context and subsequently helps in training the decoder faster.

\section{Conclusion}
\label{conclusion}

The proposed model, Extended Encoder Decoder, offers a framework to incorporate unstructured external knowledge to generate dialogue utterances. The experiments helped in understanding the need for external knowledge sources for improving learning time, and helped characterize the value of external knowledge sources. The experiments showed that external knowledge improved the learning time of the models. In future work, we aim to add more experiments with dialogue tasks that require understanding a supplementary source of knowledge to solve the task. Also, we plan to look at specialized tasks that naturally evaluate the influence of external knowledge, to help the model to generate diverse responses.
\section*{Acknowledgements}
The authors gratefully acknowledge
financial support for portions of this work by
the Samsung Advanced Institute of Technology
(SAIT) and the Natural Sciences and Engineering
Research Council of Canada (NSERC).
\bibliographystyle{acl_natbib_nourl}
\bibliography{emnlp2018}

\begin{thebibliography}{25}
\expandafter\ifx\csname natexlab\endcsname\relax\def\natexlab#1{#1}\fi

\bibitem[{Bordes et~al.(2016)Bordes, Boureau, and Weston}]{bordes2016learning}
Antoine Bordes, Y-Lan Boureau, and Jason Weston. 2016.
\newblock Learning end-to-end goal-oriented dialog.
\newblock In \emph{arXiv preprint}.

\bibitem[{Carlson et~al.(2010)Carlson, Betteridge, Kisiel, and
  Settles}]{carlson2010toward}
Andrew Carlson, Justin Betteridge, Bryan Kisiel, and Burr Settles. 2010.
\newblock Toward an architecture for never-ending language learning.
\newblock In \emph{AAAI}.

\bibitem[{Cho et~al.(2014)Cho, Van~Merri{\"e}nboer, Bahdanau, and
  Bengio}]{cho2014gru}
Kyunghyun Cho, Bart Van~Merri{\"e}nboer, Dzmitry Bahdanau, and Yoshua Bengio.
  2014.
\newblock On the properties of neural machine translation: Encoder-decoder
  approaches.
\newblock In \emph{arXiv preprint}.

\bibitem[{Dahl et~al.(2012)Dahl, Yu, Deng, and Acero}]{speechrec}
George~E Dahl, Dong Yu, Li~Deng, and Alex Acero. 2012.
\newblock Context-dependent pre-trained deep neural networks for
  large-vocabulary speech recognition.
\newblock In \emph{IEEE Transactions on audio, speech, and language
  processing}.

\bibitem[{Ghazvininejad et~al.(2017)Ghazvininejad, Brockett, Chang, Dolan, Gao,
  Yih, and Galley}]{ghazvininejad2017knowledge}
Marjan Ghazvininejad, Chris Brockett, Ming-Wei Chang, Bill Dolan, Jianfeng Gao,
  Wen-tau Yih, and Michel Galley. 2017.
\newblock A knowledge-grounded neural conversation model.
\newblock In \emph{arXiv preprint}.

\bibitem[{Graves et~al.(2014)Graves, Wayne, and Danihelka}]{graves2014neural}
Alex Graves, Greg Wayne, and Ivo Danihelka. 2014.
\newblock Neural turing machines.
\newblock In \emph{arXiv preprint}.

\bibitem[{Gulcehre et~al.(2018)Gulcehre, Chandar, Cho, and
  Bengio}]{gulcehre2018dynamic}
Caglar Gulcehre, Sarath Chandar, Kyunghyun Cho, and Yoshua Bengio. 2018.
\newblock Dynamic neural turing machine with continuous and discrete addressing
  schemes.
\newblock In \emph{Neural computation}. MIT Press.

\bibitem[{Guu et~al.(2017)Guu, Hashimoto, Oren, and Liang}]{guu2017generating}
Kelvin Guu, Tatsunori~B Hashimoto, Yonatan Oren, and Percy Liang. 2017.
\newblock Generating sentences by editing prototypes.
\newblock In \emph{arXiv preprint}.

\bibitem[{Hochreiter and Schmidhuber(1997)}]{lstm}
Sepp Hochreiter and J{\"u}rgen Schmidhuber. 1997.
\newblock Long short-term memory.
\newblock In \emph{Neural computation}.

\bibitem[{Katz(1987)}]{katzNgram}
Slava Katz. 1987.
\newblock Estimation of probabilities from sparse data for the language model
  component of a speech recognizer.
\newblock In \emph{IEEE transactions on acoustics, speech, and signal
  processing}.

\bibitem[{Levin and Pieraccini(1997)}]{pipelinedialogue}
Esther Levin and Roberto Pieraccini. 1997.
\newblock A stochastic model of computer-human interaction for learning
  dialogue strategies.
\newblock In \emph{Fifth European Conference on Speech Communication and
  Technology}.

\bibitem[{Li et~al.(2017)Li, Monroe, Shi, Ritter, and
  Jurafsky}]{li2017adversarial}
Jiwei Li, Will Monroe, Tianlin Shi, Alan Ritter, and Dan Jurafsky. 2017.
\newblock Adversarial learning for neural dialogue generation.
\newblock In \emph{arXiv preprint}.

\bibitem[{Long et~al.(2017)Long, Wang, Xu, Wang, Wang, and Wang}]{longenhanced}
Yinong Long, Jianan Wang, Zhen Xu, Zongsheng Wang, Baoxun Wang, and Zhuoran
  Wang. 2017.
\newblock A knowledge enhanced generative conversational service agent.
\newblock In \emph{the 6th Dialog System Technology Challenges (DSTC6)
  Workshop}.

\bibitem[{Lowe et~al.(2015)Lowe, Pow, Serban, Charlin, and
  Pineau}]{lowe2015incorporating}
Ryan Lowe, Nissan Pow, Iulian Serban, Laurent Charlin, and Joelle Pineau. 2015.
\newblock Incorporating unstructured textual knowledge sources into neural
  dialogue systems.
\newblock In \emph{Neural Information Processing Systems Workshop on Machine
  Learning for Spoken Language Understanding}.

\bibitem[{Rumelhart et~al.(1988)Rumelhart, Hinton, Williams et~al.}]{rumelhart}
David~E Rumelhart, Geoffrey~E Hinton, Ronald~J Williams, et~al. 1988.
\newblock Learning representations by back-propagating errors.
\newblock In \emph{Cognitive modeling}.

\bibitem[{Scheepers(2017)}]{scheepers2017compositionality}
Thijs Scheepers. 2017.
\newblock Improving the compositionality of word embeddings.
\newblock Master's thesis, Universiteit van Amsterdam.

\bibitem[{Schuster and Paliwal(1997)}]{bidirectional}
Mike Schuster and Kuldip~K Paliwal. 1997.
\newblock Bidirectional recurrent neural networks.
\newblock In \emph{IEEE Transactions on Signal Processing}.

\bibitem[{Serban et~al.(2015)Serban, Lowe, Charlin, and
  Pineau}]{serban2015survey}
Iulian~Vlad Serban, Ryan Lowe, Laurent Charlin, and Joelle Pineau. 2015.
\newblock A survey of available corpora for building data-driven dialogue
  systems.
\newblock In \emph{arXiv preprint}.

\bibitem[{Serban et~al.(2016{\natexlab{a}})Serban, Sordoni, Bengio, Courville,
  and Pineau}]{serban2016building}
Iulian~Vlad Serban, Alessandro Sordoni, Yoshua Bengio, Aaron~C Courville, and
  Joelle Pineau. 2016{\natexlab{a}}.
\newblock Building end-to-end dialogue systems using generative hierarchical
  neural network models.
\newblock In \emph{AAAI}.

\bibitem[{Serban et~al.(2016{\natexlab{b}})Serban, Sordoni, Lowe, Charlin,
  Pineau, Courville, and Bengio}]{serban2016hierarchical}
Iulian~Vlad Serban, Alessandro Sordoni, Ryan Lowe, Laurent Charlin, Joelle
  Pineau, Aaron Courville, and Yoshua Bengio. 2016{\natexlab{b}}.
\newblock A hierarchical latent variable encoder-decoder model for generating
  dialogues.
\newblock In \emph{arXiv preprint}.

\bibitem[{Sukhbaatar et~al.(2015)Sukhbaatar, Weston, Fergus
  et~al.}]{sukhbaatar2015end}
Sainbayar Sukhbaatar, Jason Weston, Rob Fergus, et~al. 2015.
\newblock End-to-end memory networks.
\newblock In \emph{Advances in neural information processing systems}.

\bibitem[{Sutskever et~al.(2014)Sutskever, Vinyals, and Le}]{seq2seq}
Ilya Sutskever, Oriol Vinyals, and Quoc~V. Le. 2014.
\newblock Sequence to sequence learning with neural networks.

\bibitem[{Vinyals and Le(2015)}]{neural_conv}
Oriol Vinyals and Quoc~V. Le. 2015.
\newblock A neural conversational model.

\bibitem[{Yang et~al.(2013)Yang, Zhang, Yu, and Yu}]{foursquare}
Dingqi Yang, Daqing Zhang, Zhiyong Yu, and Zhiwen Yu. 2013.
\newblock Fine-grained preference-aware location search leveraging crowdsourced
  digital footprints from lbsns.
\newblock In \emph{Proceedings of the 2013 ACM international joint conference
  on Pervasive and ubiquitous computing}.

\bibitem[{Young et~al.(2017)Young, Cambria, Chaturvedi, Huang, Zhou, and
  Biswas}]{young2017augmenting}
Tom Young, Erik Cambria, Iti Chaturvedi, Minlie Huang, Hao Zhou, and Subham
  Biswas. 2017.
\newblock Augmenting end-to-end dialog systems with commonsense knowledge.
\newblock In \emph{arXiv preprint}.

\end{thebibliography}
\end{document}